\documentclass{article}
\usepackage{ijcai22}

\usepackage{times}

\usepackage{soul}
\usepackage{url}
\usepackage[hidelinks]{hyperref}
\usepackage[utf8]{inputenc}
\usepackage[small]{caption}
\usepackage{graphicx}
\usepackage{booktabs}
\usepackage{bm}
\usepackage{amsmath}
\usepackage{xcolor}
\usepackage{amsfonts}

\title{Multi-Frequency-Aware Patch Adversarial Learning\\ for Neural Point Cloud Rendering}

\author{
Jay	Karhade$^1$\footnote{Equal contribution
}\and
Haiyue Zhu$^{2*}$\footnote{Corresponding author}\and
Ka-Shing Chung$^{3*}$\and
Rajesh Tripathy$^1$\and\\
Wei Lin$^2$\And
Marcelo H. Ang Jr.$^{3}$\\
\affiliations
$^1$Birla Institute of Technology and Science, Pilani\\
$^2$SIMTech, Agency for Science, Technology and Research\\
$^3$National University of Singapore
\emails
jaykarhade3@gmail.com, zhu\_haiyue@simtech.a-star.edu.sg,
chung@u.nus.edu,
tripathyrk@hyderabad.bits-pilani.ac.in,
wlin@simtech.a-star.edu.sg,
mpeangh@nus.edu.sg
}

\begin{document}

\maketitle

\begin{abstract}
We present a neural point cloud rendering pipeline through a novel multi-frequency-aware patch adversarial learning framework. The proposed approach aims to improve the rendering realness by minimizing the spectrum discrepancy between real and synthesized images, especially on the high-frequency localized sharpness information which causes image blur visually. Specifically, a patch multi-discriminator scheme is proposed for the adversarial learning, which combines both spectral domain (Fourier Transform and Discrete Wavelet Transform) discriminators as well as the spatial (RGB) domain discriminator to force the generator to capture global and local spectral distributions of the real images. The proposed multi-discriminator scheme not only helps to improve rendering realness, but also enhance the convergence speed and stability of adversarial learning. Moreover, we introduce a noise-resistant voxelisation approach by utilizing both the appearance distance and spatial distance to exclude the spatial outlier points caused by depth noise. Our entire architecture is fully differentiable and can be learned in an end-to-end fashion. Extensive experiments show that our method produces state-of-the-art results for neural point cloud rendering by a significant margin. Our source code will be made public at a later date.
\end{abstract}

\section{Introduction}

Photo-realistic rendering from point cloud has attracted increasing attention as point cloud is a well-accepted format that is widely used in various vision tasks. Novel view synthesis for given camera viewpoint allows the user an alternative choice to physical presence, which is in high demand for many applications. However, due to the inherent irregularity and discontinuity, view rendering from 3D scene involve complex graphic pipelines that include multiple pre-processing and post-processing steps. Traditional model-based rendering~\cite{Dai2017} aims to reconstruct surfaces and render on the mesh by employing the physical properties of lighting, texture, material, etc., which is generally computationally heavy. Image-based rendering~\cite{Hedman2018Deep} attempts to generate the novel view from images only, and recent point-based rendering~\cite{dai2020neural,song2020,aliev2020neural,riegler2020free} further simplifies the geometry constructions in rendering. Deep learning approaches~\cite{goodfellow2014generative,isola2018imagetoimage,zhu2020unpaired} are widely adopted in view of its superior performance in almost all kinds of image reconstruction tasks.
 
 \begin{figure}[tb!]
\includegraphics[width=8.0cm]{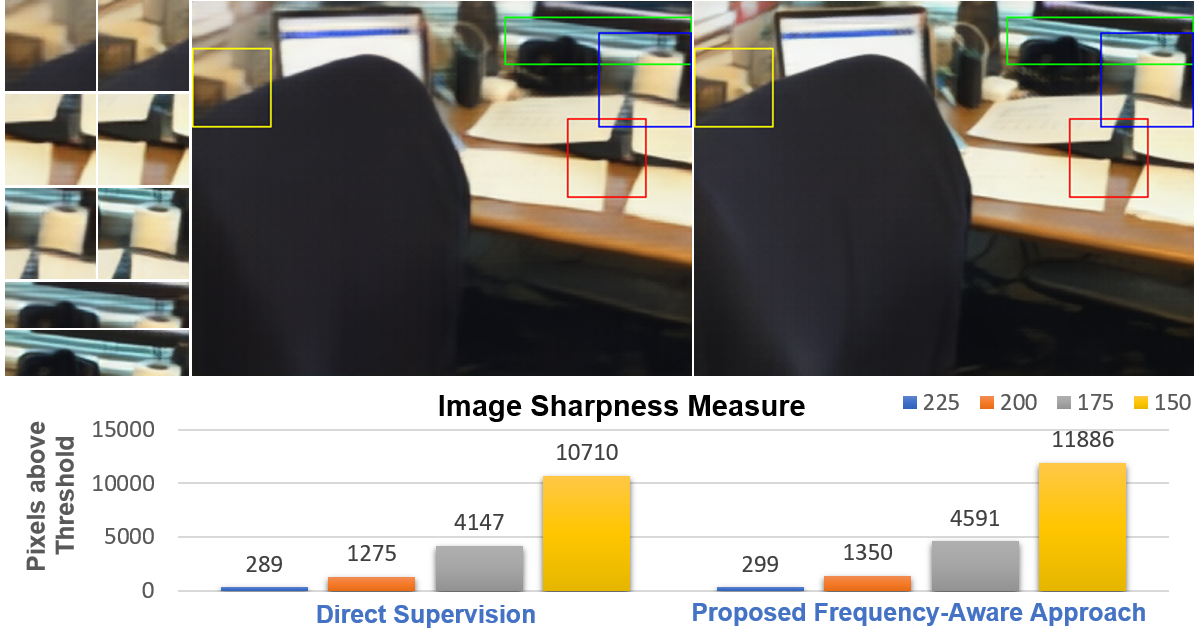}
\centering
\caption{Deep rendering network trained by direct supervision often lacks high-frequency details. We propose a multi-frequency-aware adversarial learning approach to minimize such frequency discrepancy so that to improve the sharpness and realness. The bar chart shows the sharpness comparison using the image sharpness measure proposed in \protect\cite{De2013measure}. Our method is visually also more well-defined: The glass mug, window shades, chair arm-rest appear much more defined with our method.}
\label{fig::sharpness}
\end{figure} 

To achieve photo-realistic rendering, existing neural point cloud rendering approaches mostly emphasize on the reconstruction performance in RGB (spatial) domain only while overlooking image fidelity in the spectral domain. It is also well reported that convolutional neural network based image reconstruction approaches often fail to generalize high-frequency artifacts~\cite{fritsche2019frequency,dzanic2020fourier,Giudice2021Fighting}, causing the synthesized image to suffer from a lack of high-frequency sharpness. This spectrum discrepancy issue is a performance-limiting factor that is common to image reconstruction and generation tasks as well as neural point cloud rendering tasks. Existing techniques from image generation and super-resolution make use of a Fourier regularization term~\cite{Durall2020Watch} or loss~\cite{fuoli2021fourier} on top of the Generative Adversarial Networks (GAN) to overcome the frequency discrepancy limitation for image generation and super-resolution tasks. However, as it is well-known Fourier transform lacks of spatially-localized frequency information to capture the abrupt signal changes, existing methods still inhibit the full utilization to well capitalize spectral domain discrepancy and recover high-frequency sharpness information.

Moreover, different with most image-to-image translation vision tasks, point cloud rendering requires a voxelisation projection step to extract the view-related 3D points according to the given rendering viewpoint. \cite{aliev2020neural} directly projects 3D points onto the 2D plane, which is sensitive to the spatial occlusion and noise. Multi-plane projection~\cite{dai2020neural} is proposed as a remedy by considering the spatial distance of 3D points as a measure in the feature aggregation. However, with regards to the sensor characteristics of an RGB-D device being used to produce the initial colored point cloud, it is well-known that depth channel is generally associated with more significant noise especially for those commonly-used low-cost RGB-D cameras. Therefore, we argue that the appearance (RGB) feature of each 3D point is more robust than its spatial location, which could be utilized in the aggregation for robust view rendering. 

In this work, we present a neural point cloud rendering pipeline that aims to bridge the above-mentioned gaps. A novel multi-frequency-aware patch adversarial learning scheme is proposed to supervise the point cloud rendering generation, which is specifically designed to promote the high-frequency generalization capability to minimize the local frequency discrepancy. Different from existing Fourier transform based frequency-aware adversarial learning, we explore the Discrete-Wavelet Transform (DWT) based learning as DWT is known to capture the abrupt changes of a signal more effectively, which provides additional spatially-localized high-frequency information for adversarial discrimination. Therefore, we propose a multi-discriminator strategy embedding with both Fourier domain and DWT domain discriminators as well as the spatial domain discriminator to force the generator to capture global and local spectral distributions of the real scene images. This not only helps to improve realness of localized artifacts, but also enhance the convergence speed and stability of adversarial learning. Moreover, a noise-resistant voxelisation approach is proposed by utilizing both the appearance feature distance and spatial distance for robust voxel feature aggregation. By incorporating the more stable appearance distance into feature voting, the spatial outlier points caused by depth noise can be excluded more efficiently. Our entire architecture is differentiable and can be learned in an end-to-end fashion.

To summarize, the contributions of this paper are listed as follows:
\begin{itemize}
\setlength\itemsep{-.3em}
\item We propose a multi-frequency-aware adversarial learning scheme, which utilizes the under-explored DWT domain to minimize the localized spectrum discrepancy between generated images and real images.
\item We introduce a combined feature and spatial distance based noise-resistant voxelisation approach for robust neural point cloud rendering.
\item Our approach effectively enhances the realness and sharpness of the generated images, and achieves state-of-the-art results by a significant margin.
\end{itemize}

\begin{figure*}[tb!]
\includegraphics[height=10cm, width=18cm]{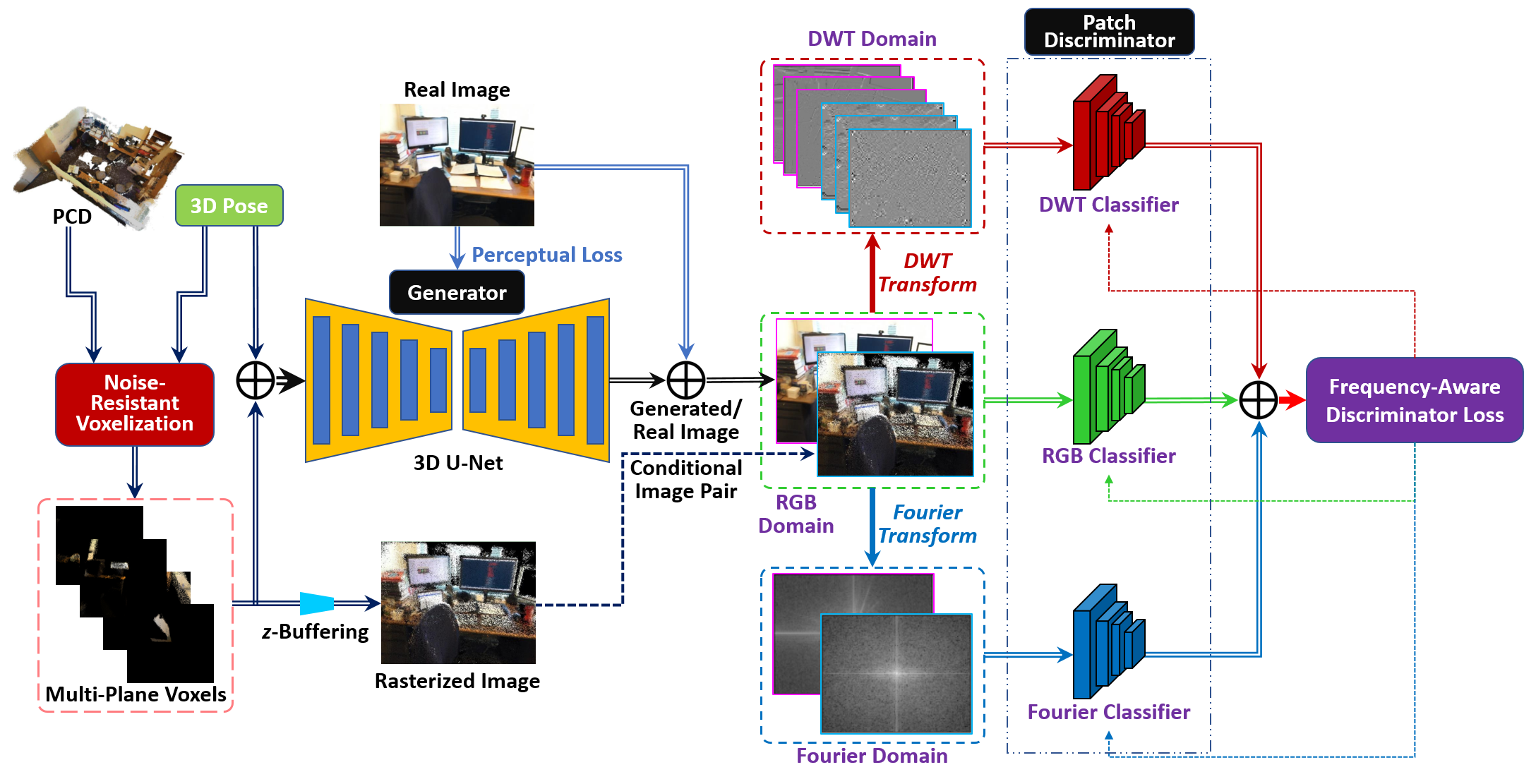}
%\vspace{-1.5cm}
\caption{Method overview: we use point cloud and 3D pose to first voxelise the visible point frustrum into multi-plane projections using the proposed Noise-Resistant Voxelisation approach. This voxelisation input is passed into the 3D U-Net generator to synthesize the view rendering image. This generator is supervised through a multi-frequency-aware patch adversarial learning scheme as well as a standard perceptual loss, where a discriminator network is trained simultaneously to differentiate between real and generated images. Uniquely, to better capture high-frequency discrepancy, we adopt a tri-discriminator network that combines both spectral domain (Fourier Transform and Discrete Wavelet Transform) discriminators as well as the spatial (RGB) domain information to minimize the frequency discrepancy.}
\label{fig::method}
\end{figure*} 

\section{Related works}
\subsection{Novel View Synthesis}
Recently, several works have been proposed for view synthesis of objects~\cite{ramirez2021unsupervised} and human faces~\cite{huang2017face}. A number of image-based novel view approaches have been proposed for scene rendering. For instance, Synsin~\cite{wiles2020synsin} generates a latent point cloud representation from a single image and produces a target image. While this is highly effective for small deviations of pose from the image, large deviations produce undesirable artifacts at the corners. Other recent approaches regarding image synthesis like Stable View Synthesis~\cite{riegler2021stable} and Free View Synthesis~\cite{riegler2020free} use a series of images to create a geometric scaffold or a mesh followed by feature aggregation of a ray at sampled positions on the mesh from different viewpoints and rendering. Another work, NeRF~\cite{mildenhall2020nerf}, makes use of a hybrid deep learning and classical volume rendering approach, and several works inspired from NeRF such as \cite{yu2021pixelnerf}, \cite{barron2021mipnerf}, \cite{martinbrualla2021nerf}, \cite{zhang2020nerf++} and \cite{tancik2020fourier} have been proposed for improved efficiency and limited image data. Our work differs from these since we use raw point clouds as input instead of images.

\subsection{Neural Point Rendering}
Deep novel view synthesis~\cite{song2020} extracts features from a point encoder layer to encode an input point set and passes it on to an image decoder followed by a refinement network. While effective for sparse point clouds, the PointNet++ backbone cannot capture local relationships effectively and requires large paired data for refinement. Neural Point Graphics (NPG)~\cite{aliev2020neural} proposes augmenting learnable neural descriptors for point clouds and rasterizes it to a 2D image for multi-scale rendering network. Using multi-plane projections significantly improves performance and recent work such as~\cite{Flynn_2019_CVPR} utilize the advantage of a layered volume. \cite{dai2020neural} extends NPG by also using multi-plane projections, which is the closest work to ours, and it is used as a baseline for comparison in our experiments.

%\subsection{Image Transformation Network}
%Deep image translation networks have shown success in numerous tasks like image translation, inpainting, and deblurring~\cite{kupyn2019deblurganv2,lu2019unsupervised} for both paired and unpaired data. Pix2Pix~\cite{isola2018imagetoimage} and CycleGAN~\cite{zhu2020unpaired} are two seminal papers for image translation. However, an image translation approach cannot be directly applied to rendering as it will fail to capture underlying depth information. By using a 3D U-Net, we overcome this limitation and our focus is to reduce the high-frequency discrepancy and promote the sharpness and realness of the rendering image in this work.

\subsection{Spectral Domain Loss}
With the variety of image generation/translation methods in the literature, image blur along with lack of sharpness or the presence of high-frequency artifacts is an observed issue which limits the visual realness. \cite{tancik2020fourier} have shown that neural networks tend to learn lower frequencies faster, and overcome this by using Fourier mapping layers prior to pass inputs to a multi-layer perceptron (MLP) network for higher perceptual quality. Recent works~\cite{Giudice2021Fighting} show that the realness degradation is partly attributable to missing high-frequency features due to the spectrum discrepancy. Furthermore, \cite{Durall2020Watch} and \cite{zhang2019detecting} recognize that GAN based models, especially those with upsampling layers, fail to generalize spectral data. To alleviate this problem, works such as \cite{Durall2020Watch} have proposed a Fourier transform as a regularization term. Arguing regularization may lead to sub-optimal performance, \cite{chen2020ssdgan} and \cite{fuoli2021fourier} show that using an additional discriminator based Fourier loss leads to better performance. \cite{Giudice2021Fighting} uses the direct-cosine transform as a better perceptual loss. \cite{liu2018multilevel} employs a Discrete Wavelet Transform (DWT) module to keep frequency information as a down-sampling module in the generator. Although it is also a DWT-based GAN approach, our motivation and approach are totally different from~\cite{liu2018multilevel} as we favour a patch multi-discriminator strategy. We aim to reduce the frequency discrepancy in both spectral (DWT and Fourier) and spatial (RGB) domains, so that the generator can generalize better on the source distribution to enhance the visual realness.

\section{Method}

\subsection{Overview}
The rendering problem studied in this work is formulated as follows. Given a scene point cloud $\bm{P}=\{\bm{p}_{1}, \bm{p}_{2}, \cdots, \bm{p}_{N}\}$ with a set of $M$ camera images taking from the same scene and their corresponding camera poses, denoted as $\bm{C}_{M}=\{\bm{I}, \bm{c}\}_{M}$, the goal of neural point cloud rendering is to learn a mapping function $f(\cdot)$ that can render a virtual image $\hat{\bm{I}}$ from the point cloud $\bm{P}$ with a random target camera pose $\bm{c} \in \mathbb{R}^{3\times4}$, i.e., 
\begin{equation}
\hat{\bm{I}} = f(\bm{P},\bm{c}),
\end{equation}
where the generated image $\hat{\bm{I}}$ is aimed to be perceptually realistic as much as possible.

Fig.~\ref{fig::method} shows our neural rendering pipeline in this work, where we propose a frequency-aware patch adversarial learning to achieve the photo-realistic rendering. Our approach learns a rendering generator $G(\cdot)$ through the adversarial learning, which takes the voxelised 3D volume with the target pose to synthesize the target image. A noise-resistant voxelisation method is developed to produce consistent 3D volumes even from noisy and irregularly sampled point clouds, and a multi-frequency-aware patch discriminator scheme is proposed in the adversarial learning to effectively capture the high frequency visual information both locally and globally through the Fourier domain and DWT domain. 

\subsection{Noise-Resistant Voxelisation}
Following~\cite{dai2020neural}, we project the visible point cloud region into multiple planes so as to avoid artifacts from occluded regions. This projection helps to correct noise interference in comparison with just using a one-plane rasterized image. For a voxel centered at $(p,h,w)$ that contains $N_v$ number of sub-voxel points with features as $f_{p,h,w}^{i}$, the voxel feature $F_{(p,h,w)}$ is calculated by aggregation of all sub-voxel points. \cite{dai2020neural} calculates a weighted average of sub-voxel points by considering the spatial distance between the vertical and horizontal planes. However, such purely spatial distance based voxelisation is sensitive to the noisy points closer to the centre, which could hamper the rendering performance. 

In contrast, the appearance feature of a point is generally more robust than the spatial location considering the sensor physical properties of RGB-D devices. Therefore, we propose a noise-resistant voxelisation that incorporates the feature distance as well to aggregate the voxel feature,
\begin{equation}
\begin{aligned}
\label{eq::Voxelization}
F_{(p,h,w)} =&\frac{{\sum_{i} w_{(p,h,w)}^{i} f_{(p,h,w)}^{i}}}{{\sum_{i} w_{(p,h,w)}^{i}}},
\end{aligned}
\end{equation}
where the blending weight $w_i$ considers both the spatial and feature distances as
\begin{equation}
\begin{aligned}
w^{i}_{(p,h,w)} &= \mu_{f}{D_{f(p,h,w)}^{i}}+\mu_{s}D^{i}_{s(p,h,w)}\\
D^{i}_{s(p,h,w)}&=(1-D_{1(p,h,w)}^i)^\alpha(1+D_{2(p,h,w)}^i)^\beta\\
D_{f(p,h,w)}^{i}&={(|f_{p,h,w}^{i}-\Bar{f}_{p,h,w}|}_1)^{-1} 
\end{aligned}
\end{equation}
where $D^{i}_{s}$ and $D^{i}_{f}$ are the spatial and feature inverse distances, respectively, $\mu_{f}$ and $\mu_{s}$ are the weights controlling their effects. $\Bar{f}_{p,h,w}$ is the average of point color features for points in voxel $({p,h,w})$, $\Bar{f}_{p,h,w}=\sum_{i=0}^{N}{f_{p,h,w}^{i}}/N$. $\alpha$ and $\beta$ are hyper-parameters to control the blending weight; $D_1$ and $D_2$ are the distances of the point from center of voxel and minimum depth point for a particular voxel, respectively. Overall, the voxelisation can be represented as
\begin{equation}
\bm{F}=\mathcal{V}(\bm{P},\bm{c}),
\end{equation}
where $\mathcal{V}$ denotes the proposed noise-resistant voxelisation operation.

\subsection{Patch Adversarial Learning}
For the rendering network, we employ the adversarial  scheme to learn the generator mapping from the voxelisation input $\bm{F}$ to the synthesized image rendering $\bm{\hat{I}}$, i.e.,
$\hat{\bm{I}}=G(\bm{F})$, which is a simple 3D U-Net.
The adversarial counterpart is a conditional patch discriminator array $D(\cdot)$, which takes the input patch pairs to distinguish whether it is real camera image or synthesized image. For the conditional pairs, we perform a $z$-buffer rasterization of the multi-plane voxelisation and concatenate it with either the synthesized or real image as the input of $D(\cdot)$. Inspired by PatchGAN~\cite{isola2018imagetoimage}, the discriminator adopts the patch approach to output $N_{p}\times N_{p}$ scores, which penalizes the discrimination for each receptive region to encourage the attention on high-frequency local information. The patch loss is especially effective for our DWT frequency discriminator since the spectral features are local and can be better exploited to distinguish fake/real images to improve the generator performance.

\subsection{Frequency-Aware Multiple Discriminator}

One major limitation of traditional view rendering, many of which optimize either an $L_2$ or $L_1$ loss, is that it often only captures the low frequency visual details, leading to the sharpness degradation on high frequency abrupt changes. Our adversarial scheme is designed to be frequency-aware to effectively capture and generalize high-frequency visual information. Our network uses a multi-discrimination strategy that combines the RGB domain with frequency domain to improve image synthesis quality indirectly. The strategy is achieved by a Fourier discriminator that encourages the global generalization of high frequencies, as well as a DWT discriminator, which locally differentiates the frequency discrepancy in the image. As the DWT and Fourier discriminators spotlight contrasting aspects of discrepancy on frequency domain, the two work in tandem to drive the generator to produce a more photo-realistic rendering result.

Our multi-frequency-aware discriminator scheme is formed as
\begin{equation}
\label{eq::multi-discriminator}
    D(\cdot)=D_{RGB}(\cdot)+D_{Fourier}(\cdot)+D_{DWT}(\cdot),
\end{equation}
where $D_{RGB}(\cdot)$ is a patch discriminator on raw RGB domain, which takes the input pair $\{\bm{I}_{r},\bm{I}\backslash\bm{\hat{I}}\}$ formed by the rasterized image $\bm{I}_{r}$ with either the real image $\bm{I}$ or generated image $\bm{\hat{I}}$. $D_{Fourier}(\cdot)$ is a frequency discriminator on Fourier domain, which takes the Fourier transformed input pair $\{\mathcal{F}(\bm{I}_{r}),\mathcal{F}(\bm{I})\backslash\mathcal{F}(\bm{\hat{I}})\}$, where $\mathcal{F}(\cdot)$ denotes the Fourier transform operation. It is well-known that Fourier transform cannot well capture the abrupt changes of a signal which is a limitation for measuring high-frequency artifacts. To solve this issue, we propose using a DWT discriminator $D_{DWT}$ in addition to $D_{Fourier}$ in~\eqref{eq::multi-discriminator}. The DWT is aiming for those perceptually-important localized frequency information, and we take three sub-bands, i.e., HL (vertical), LH (horizontal), and HH (diagonal)  features from the DWT, as input to $D_{DWT}$, denoted as $\{\mathcal{HL}(\bm{I}_{r}),\mathcal{LH}(\bm{I}_{r}),\mathcal{HH}(\bm{I}_{r}),\mathcal{HL}(\bm{I}\backslash\bm{\hat{I}}),\mathcal{LH}(\bm{I}\backslash\bm{\hat{I}}),\mathcal{HH}(\bm{I}\backslash\bm{\hat{I}}\}$,
where $\mathcal{HL}$, $\mathcal{LH}$, and $\mathcal{HH}$ denotes the respective components.

\begin{table*}[htb!]
    \centering
    \caption{Performance Comparison on Scannet and Matterport 3D}
\label{tbl::Datasetcomparison}
\begin{tabular}{ccccccc}
\toprule
&\multicolumn{3}{c}{ScanNet} &\multicolumn{3}{c}{Matterport 3D} \\ 
\midrule
\textbf{Methods} & \textbf{SSIM} & \textbf{PSNR} & \textbf{LPIPS} & \textbf{SSIM} & \textbf{PSNR} & \textbf{LPIPS}  \\ \hline
Pix2Pix~\cite{isola2018imagetoimage}     &0.731 & 19.247 &0.429    &0.530 &  14.964 & 0.675         \\ 
NPG~\cite{aliev2020neural}  &0.84   &22.911    &0.245   &0.622  &22.911    & 0.597  \\ 
MPP~\cite{dai2020neural}  &0.835   &22.813    &0.234   &0.649   &18.09    & 0.534  \\ 
\textbf{Ours}   &\textbf{0.871} &\textbf{32.7}   &\textbf{0.1012}   &\textbf{0.672}    &\textbf{25.95}   &\textbf{0.24} \\
\bottomrule
\end{tabular}
\end{table*}

\subsection{Loss Functions}

In the proposed frequency-aware patch adversarial learning,
each discriminator is supervised using least square patch loss,
\begin{equation}
\begin{aligned}
\label{eq::Dloss}
\mathcal{L}_{D_{\pi}} &=
\mathbb{E}_{\bm{I}} \sum_{i=0}^{N_{p}-1}\sum_{j=0}^{N_{p}-1}(D_{\pi}(\bm{I})_{i, j}-a)^{2}\\
&\quad+\mathbb{E}_{\bm{F}} \sum_{i=0}^{N_{p}-1}\sum_{j=0}^{N_{p}-1}(D_{\pi}(G(\bm{F}))_{i, j}-b )^{2}\\
\end{aligned}
\end{equation}
where $\pi\in\Pi$ and $\Pi=[RGB,Fourier,DWT]$. $a$ and $b$ are the labels for fake data and real data, and $D(\cdot)_{i, j}$ denotes the $i, j$-th patch prediction. As a result, the overall objective for the discriminator is
\begin{equation}
\begin{aligned}
\min_{D}\mathcal{L}(D) &=\min_{D_{\pi}}\sum_{\pi\in\Pi}\mathcal{L}(D_{\pi}).
\end{aligned}
\end{equation}

For the generator, we use a perceptual loss~\cite{johnson2016perceptual} for full supervision in addition to the discriminators,
\begin{equation}
\mathcal{L}_{percpt}(G) = ||G(\bm{F})-\bm{I}||_1 + \sum_{l}\lambda_l||\phi_l(G(\bm{F}))-\phi_l(\bm{I})||_1
\end{equation}
where $\phi_l(\cdot)$ is $l$-th layer output of pre-trained VGG-19 and $\lambda_l$ is a controlling weight. Therefore, the overall objective for the generator is
\begin{equation}
\begin{aligned}
\label{eq::Gloss}
\min_{G}\mathcal{L}(G) &=\min_{G}\Big\{\mathcal{L}_{percpt}(G)\\
&\quad+\mathbb{E}_{\bm{F}}\sum_{\pi\in\Pi}\sum_{i=0}^{N_{p}-1}\sum_{j=0}^{N_{p}-1}(D_{\pi}(G(\bm{F}))_{i, j}-c )^{2}\Big\},\\ 
\end{aligned}
\end{equation}
where $c$ denotes the value that $G$ wants $D$ to believe for fake data following~\cite{mao2017least}.

\section{Implementation Details}
% \hl{put more details: what is the network structure of generator and discriminator? Residual Network? Three discriminators all same structure? input size? How many patch?}

% \begin{table*}[ht!]
% \centering
% \caption{Model Statistics}
% \label{model stats}
% \begin{tabular}{|c|c|c|c|}
% \hline
% \textbf{Type} & \textbf{Number of parameters} & \textbf{Computational Complexity}  &\textbf{Forward Pass(s)} \\ \hline
% U-Net Generator     &2.21M &283.98GMac &0.184            \\ \hline
% Spatial Discriminator  &2.77M   &2.01 GMac &0.0011        \\ \hline
% DWT Discriminator        &2.77M  &2.05 GMac &0.006       \\ \hline
% Fourier Discriminator   &2.76 M &1.93GMac &0.094        \\ \hline
% \end{tabular}
% \end{table*}

In our implementation, the generator is a simple 3D U-Net as per~\cite{dai2020neural} for fair comparison. It is also noted that a higher performing generator structure, such as 3D Residual U-Net, may further improve the results. We use three discriminators, i.e., one for spatial domain, one for Fourier domain, and one for DWT domain. Due to memory limitation, we crop the input into $240\times320$.
The input to spatial discriminator is a concatenation of the RGB generated/real image and the rasterized image with shape of $240\times320\times6$. The input to Fourier discriminator is the magnitude channels of the Fourier transformation of the above concatenation with shape of $240\times320\times2$, where both images are first converted into grayscale.
For DWT discriminator, we perform a DWT transform on the grayscale concatenation, and use the
$\mathcal{HL}$, $\mathcal{LH}$, and $\mathcal{HH}$ channels of decomposed frequency information, leading to input shape of $120\times160\times6$. All three discriminators use five convolution layers, with the output patch $16\times16$ for both spatial and Fourier discriminators and $10\times10$ for DWT discriminator. The overall loss for supervising the adversarial learning is~\eqref{eq::Dloss} and~\eqref{eq::Gloss}.

%We first use the RGB-D scans of the datasets to form a point-cloud volume. To reduce computational requirements, we simplify the point-cloud by grid sub-sampling of the entire point cloud. This is followed by voxelization and feature aggregation for particular poses for input to the the proposed model. 

For the voxelisation, we follow the setting in~\cite{dai2020neural} to use 32 planes for splitting the point cloud volume. The feature aggregating is conducted by using~\eqref{eq::Voxelization}, where the weighted average of hyperparameters $\mu_{f}$ and $\mu_{s}$ are chosen as 0.25 and 0.75, respectively. Furthermore, $\alpha$ and $\beta$ are both chosen as 1 for the spatial distance aggregation. As the original ScanNet and Matterport 3D dataset image sizes are $1296\times968$ and $1280\times1024$, respectively, we re-scale them in voxelisation to $640\times480$ and $640\times512$ respectively before performing a random crop into $240\times320$ for training. For each scene, we train our network for 64 epochs, which requires about 16 hours on average on an RTX 2080 Ti. We adopt Adam optimizer which decays its learning rate from 0.002 to 0.001 after 25 epochs.

\section{Experiments}
%We perform extensive experiments on two public datasets ScanNet~\cite{dai2017scannet} and Matterport~\cite{chang2017matterport3d} to evaluate our performance and compare with similar methods measured on the same datasets to demonstrate superiority. %Furthermore, we analyze the model statistics of our proposed method against baselines to see the time complexities which has been reported in Table \ref{model stats}.

\subsection{Datasets}
We perform extensive experiments on two public datasets ScanNet~\cite{dai2017scannet} and Matterport 3D~\cite{chang2017matterport3d} to evaluate our performance. The ScanNet dataset contains RGB-D scans in over 1500 environments, and we use the RGB-D scans and the 3D camera pose annotations for registering the point cloud. The Matterport 3D dataset is much more challenging with larger spread-out scenes, along with very large variation of poses.
We pre-process both datasets following the same setting in~\cite{dai2020neural}.

\begin{figure*}[tb!]
\includegraphics[height=10cm, width=17.5cm]{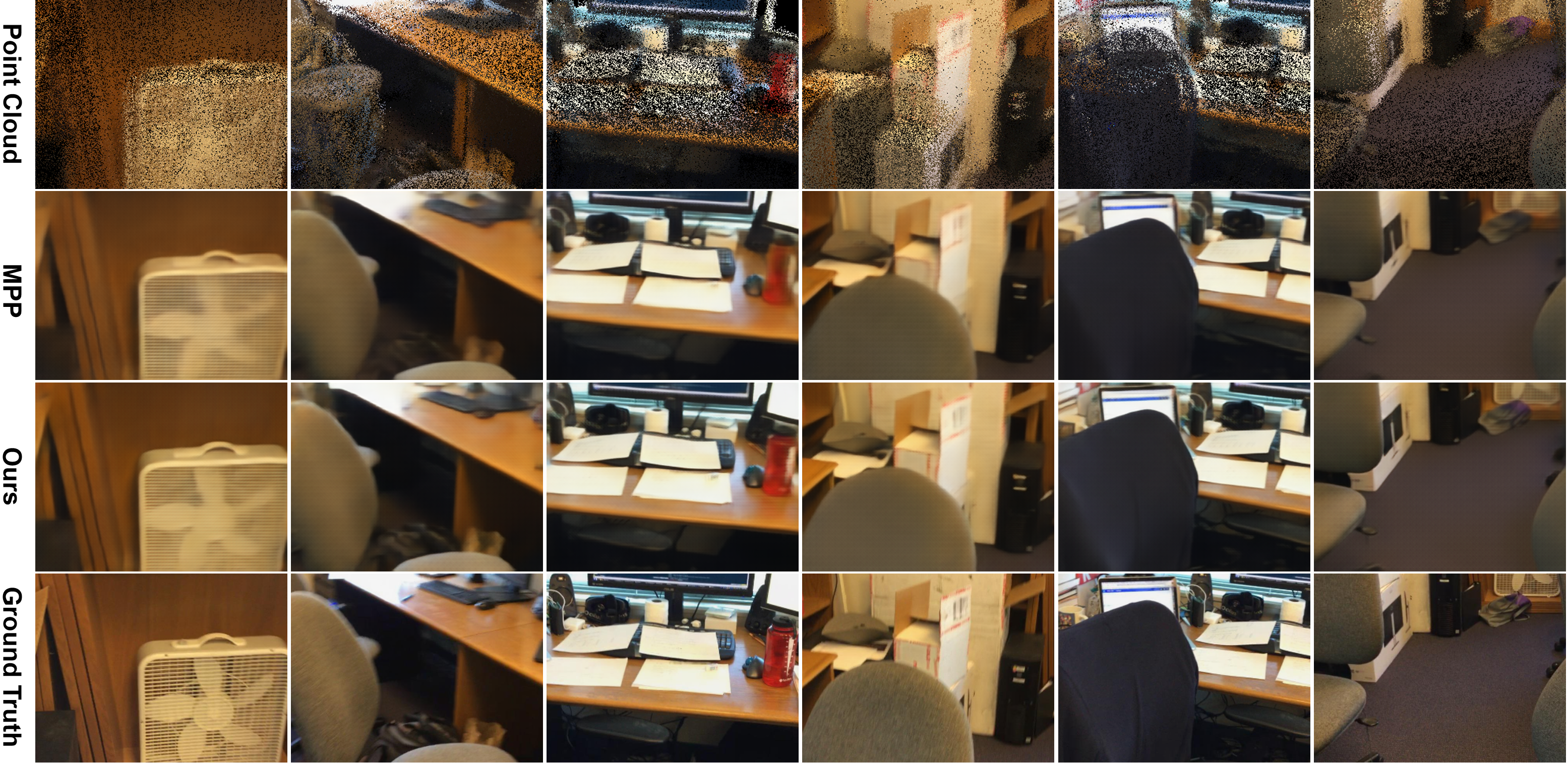}
%\vspace{-1.5cm}
\centering
\caption{Visualization of rendering results on a ScanNet scene. Objects such as the fan, chair, and water bottle have more clearly defined edges in our rendered image. Blurring artifacts such as the red colored tape on the cardboard box (fourth image from right) are also well recovered by our model. }
\label{fig::comparison}
\end{figure*} 

%The ScanNet dataset contains RGBD scans over 1500 environments, and we use the RGB-D scans and the 3D camera pose annotations for registering the point-cloud following~\cite{dai2020neural}. We reduce the size of the point-cloud in 2 methods, first by random sampling of the original scan images, and sub-sampling the point cloud via grid voxelization. With this, we are able to reduce size of a point cloud from 1.3 gigabytes to just under 200 megabytes which helps in faster feature aggregation with reduced points. The Matterport dataset is much more challenging than the ScanNet dataset, with larger spread out scenes, along with very large variation of poses which is tougher for rendering as is visible from Table \ref{Dataset comparison}. With a sub-sampling strategy of using only 1 in 30 image pixels and grid-subsampling, we reduce the point cloud size from 3gigabytes to ~ 1.3 gigabytes.

\subsection{Rendering Evaluation}

To establish the benchmarking, we consider three methods that capture ideas of image translation and point rendering techniques. Our first baseline is the well-known standard Pix2Pix network~\cite{isola2018imagetoimage} (denoted as Pix2Pix), which use a classical $z$-buffer rasterized image as input, and the translation task is to convert it to a realistic view. While this potentially interprets the distribution of noise and sparsity of point-clouds, it is unable to capture any information about pose and/or scene depth. Our next two baselines are Neural Point-based Graphic~\cite{aliev2020neural} (denoted as NPG) and Multi-Plane Projection Rendering~\cite{dai2020neural} (denoted as MPP), which capture pose information and depth information by using learnable point descriptors with multi-plane information.

To evaluate the performance of different approaches, we compare the quantitative evaluations for our method with all three baselines mentioned above. We adopt three performance metrics same as~\cite{dai2020neural}, namely Structural Similarity Index (SSIM), Peak-Signal to Noise Ratio (PSNR), and Learned Perceptual Image Patch Similarity (LPIPS). The evaluation results are provided in Table~\ref{tbl::Datasetcomparison}. It can be observed that for ScanNet, Pix2Pix leads to sub-optimal results with SSIM of only 0.73, and both of NPG and MPP lead to SSIM around 0.83$\sim$0.84. Our proposed method significantly increases the SSIM to 0.871, and similarly attains the best PSNR and LPIPS scores. We attribute this significant improvement to the combination of spectral and spatial domain losses, which leads to better perceptual quality whilst improving the Signal to Noise Ratio. As a result, our approach generates sharper and more realistic images. Fig.~\ref{fig::comparison} shows some comparison of visualization results on ScanNet dataset. Our rendering is able to define sharper object boundaries and inpaint colored features in areas missed out by the traditional rendering approach, making our synthesized image closer to the ground truth. Similar outperformance trend can be observed for Matterport 3D dataset as shown in Table~\ref{tbl::Datasetcomparison}.

% In addition to offering stability and coherence,  we are also able to produce much sharper images that are closer to the ground truth.

% Furthermore, to test the improvement in frequency information contained in the synthesized images, we calculate a Fourier transform of the images and analyze the spectral spread of data by calculating the spectral spread of data and measuring the length of principal component eigen-vectors.

% Table \ref{Dataset comparison} provides a quantiative comparison of our network. It is visible that our network obtains superior SSIM index of 0.891 which is 0.05 more than the previous benchmark. Furthermore, we can observe that our network shows competitive improvements in LPIPS and PSNR scores.

\begin{table}[ht!]
\centering
\caption{Ablation Study with Different Discriminator Module}
\label{tbl::Discriminatorcomparison}
\begin{tabular}{lccc}
\toprule
\textbf{Discriminator Type} & \textbf{SSIM} & \textbf{PSNR} & \textbf{LPIPS} \\ 
\midrule
No Discriminator     &0.835   &22.813    &0.234   \\ 
Spatial only  &0.854   &32.37    &0.1147 \\ 
Fourier only  &0.859   &31.15    &0.1042 \\ 
Spatial+Fourier &0.864  &31.56   &\textbf{0.0985} \\ 
Spatial+DWT  &0.866 &31.91   &0.1010    \\ 
Spatial+Fourier+DWT   &\textbf{0.871} &\textbf{32.7}   &{0.1012}  \\ 
\bottomrule

\end{tabular}
\end{table}

% From \ref{Dataset comparison}, we can see that our proposed method achieves superior quantitative scores by a considerable margin in all considered metrics from prior reported values of baselines in \cite{dai2020neural}. Our method achieves improvements of ~(0.04-0.05) in SSIM that corresponds to about 6.7\% enhancement. Furthermore, it is observed that our method generates massive improvement in LPIPS and PSNR metrics. We observe a 45\% improvement in PSNR quality and a 62.6\% in LPIPS quality. 

\subsection{Ablation Studies - Discriminator Modules}
Since our multi-discriminator adversarial learning employs spatial, Fourier, and DWT discriminators, we conduct ablation studies on different combinations of the multi-discriminator strategy using ScanNet dataset. Table \ref{tbl::Discriminatorcomparison} summarizes the results and demonstrates the importance of incorporating all three domains.  We observe a drop in performance in the absence of one of the spectral discriminators, and achieve the most significant improvement from using a combination of the three discriminators. 

\subsection{Ablation Studies - Noise Resistant Voxelization}
We also prove empirically  that factoring in color features during voxelization and aggregation helps in noise reduction of the projected input to the U-Net for both large and small point-clouds. The hyper-parameters $\mu_{f}$ and $\mu_{s}$, $\alpha$ and $\beta$ are kept in an equal ratio of $1$:$1$ to have an equal influence of both position and color features. 

We first synthesize a noisy point cloud (having twice the number of points compared to the original cloud) by sampling random points from a gaussian distribution centered around each point present in the original point cloud volume(normalized using max-normalization) with varying degrees of standard deviation. We further degrade the noisy point cloud by also having random color features sampled by a gaussian distribution centered at $128$ and having a standard deviation of $100$ (to nearly cover the $0$-$255$ intensity values for each color channel).

\begin{figure}[tb!]
\includegraphics[width=8.0cm]{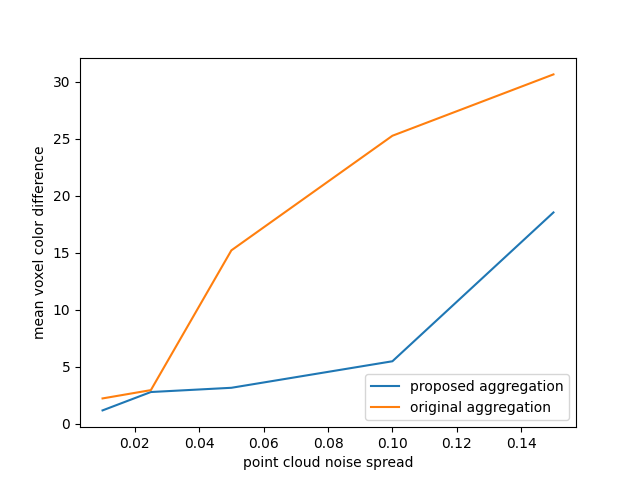}
\centering
%\vspace{-1.5cm}
\caption{We prove our method is more stable than previous methods by plotting the voxel color difference between the original voxelized point cloud and varying degrees of noisy point clouds}
\label{fig::voxelization}
\end{figure}

We then calculate the mean difference for between the aggregated colors for each voxel and its corresponding color in the original point cloud. Figure \ref{fig::voxelization} shows our proposed method consistently stays much lower than the previous method that only considers distances.

%\textcolor{red}{ZHY: No need explain so detail as people can see table. We can observe a steady improvement with adding of discriminator modules from 0.835(where we don't use any discriminator) up to 0.881(where we use a combination of all 3 discriminators). It is observed that Replacing Fourier discriminator with the DWT discriminator gives better results and raises the SSIM Index from 0.858 to 0.864. Furthermore, from experiments, we observed a stronger convergence in the training plot when using DWT discriminator. Significant improvements are reported in PSNR values on using the multi-frequency network, which indicates our network effectively captures high-frequency information and improves quality of the image. The PSNR values progressively increase from 22.813 to 33.2 on using our network, with the DWT+Spatial module giving 32.8, which is greater than using a Fourier+Spatial module. A potential interpretation of higher values when using DWT module is its ability to localize frequency features and improve accordingly.}

\section{Conclusion}

In this paper, we present a multi-frequency-aware adversarial learning scheme to achieve neural point cloud rendering, which is realized by a tri-discriminator scheme from RGB, Fourier, and DWT domains. While Fourier transform has been shown effective to regularize frequency-aware learning, it lacks sufficient generalization capabilities regarding localized high-frequency abrupt features. We combine Fourier and DWT domains with the spatial domain to achieve high fidelity and photorealistic rendering for novel view synthesis. In addition, we also introduce a noise-resistant voxelisation to reduce the impact of spatial outliers. Our model outperforms existing baselines and achieve the state-of-the-art performance on the ScanNet and Matterport 3D datasets with a significant margin.

\bibliographystyle{named}
\bibliography{ijcai22}

\end{document}